\documentclass[sigconf]{acmart}

\usepackage{graphicx}
\usepackage{amsmath}

\usepackage{amssymb}

\usepackage{booktabs}
\usepackage{wrapfig}
\usepackage{multirow}
\usepackage{diagbox}
\usepackage{threeparttable}
\usepackage{balance}
\usepackage{url}

\AtBeginDocument{%
  }

\copyrightyear{2023} 
\acmYear{2023} 
\setcopyright{acmlicensed}
\acmConference[MM '23] {Proceedings of the 31st ACM International Conference on Multimedia}{October 29--November 3, 2023}{Ottawa, ON, Canada.}
\acmBooktitle{Proceedings of the 31st ACM International Conference on Multimedia (MM '23), October 29--November 3, 2023, Ottawa, ON, Canada}
\acmPrice{15.00}
\acmISBN{979-8-4007-0108-5/23/10} 
\acmDOI{10.1145/3581783.3612465}

\begin{document}

\title{ReCo: A Dataset for Residential Community Layout Planning}
% \title{ReCo: Benchmarking Residential Community Layout Planning Generation}

\author{Xi Chen}
\email{x\_chen21@m.fudan.edu.cn}
\orcid{0009-0003-6156-4811}
\author{Yun Xiong}
\email{yunx@fudan.edu.cn}
\orcid{0000-0002-8575-5415}
\affiliation{%
  \institution{
Shanghai Key Laboratory of Data Science, \\School of Computer Science, Fudan University}
%  \streetaddress{P.O. Box 1212}
  \city{Shanghai}
  %\state{Shanghai}
  \country{China}
}

\author{Siqi Wang}
\email{siqi\_wang@tongji.edu.cn}
\orcid{0000-0003-0250-4619}
\author{Haofen Wang}
\authornote{Haofen Wang is the corresponding author.}
\email{carter.whfcarter@gmail.com}
\orcid{0000-0003-3018-3824}
\affiliation{%
  \institution{College of Design and Innovation, \\Tongji University}
%  \streetaddress{P.O. Box 1212}
  \city{Shanghai}
  % \state{Shanghai}
  \country{China}
}

\author{Tao Sheng}
\email{tsheng16@fudan.edu.cn}
\orcid{0000-0001-6744-9126}
\author{Yao Zhang}
\email{yaozhang@fudan.edu.cn}
\orcid{0000-0003-1481-8826}
\affiliation{%
  \institution{
Shanghai Key Laboratory of Data Science, \\School of Computer Science, Fudan University}
%  \streetaddress{P.O. Box 1212}
  \city{Shanghai}
  % \state{Shanghai}
  \country{China}
}

\author{Yu Ye}
\email{yye@tongji.edu.cn}
\orcid{0000-0003-3402-5786}
\affiliation{%
 \institution{College of Architecture and Urban Planning, \\Tongji University}
 \city{Shanghai}
 % \state{Shanghai}
 \country{China}}
 
\renewcommand{\shortauthors}{Xi Chen et al.}

\begin{abstract}
 Layout planning is centrally important in the field of architecture and urban design. Among the various basic units carrying urban functions, residential community plays a vital part for supporting human life. Therefore, the layout planning of residential community has always been of concern, and has attracted particular attention since the advent of deep learning that facilitates the automated layout generation and spatial pattern recognition. However, the research circles generally suffer from the insufficiency of residential community layout benchmark or high-quality datasets, which hampers the future exploration of data-driven methods for residential community layout planning. The lack of datasets is largely due to the difficulties of large-scale real-world residential data acquisition and long-term expert screening. In order to address the issues and advance a benchmark dataset for various intelligent spatial design and analysis applications in the development of smart city, we introduce \textbf{Re}sidential \textbf{Co}mmunity Layout Planning (\textbf{ReCo}) Dataset, which is the first and largest open-source vector dataset related to real-world community to date. ReCo Dataset is presented in multiple data formats with 37,646 residential community layout plans, covering 598,728 residential buildings with height information. ReCo can be conveniently adapted for residential community layout related urban design tasks, e.g., generative layout design, morphological pattern recognition and spatial evaluation.
 To validate the utility of ReCo in automated residential community layout planning, two Generative Adversarial Network (GAN) based generative models are further applied to the dataset. We expect ReCo Dataset to inspire more creative and practical work in intelligent design and beyond. The ReCo Dataset is published at: \url{https://www.kaggle.com/fdudsde/reco-dataset} and related code can be found at: \url{https://github.com/FDUDSDE/ReCo-Dataset}.
\end{abstract}



\begin{teaserfigure}
\centering
\includegraphics[width=1\linewidth]{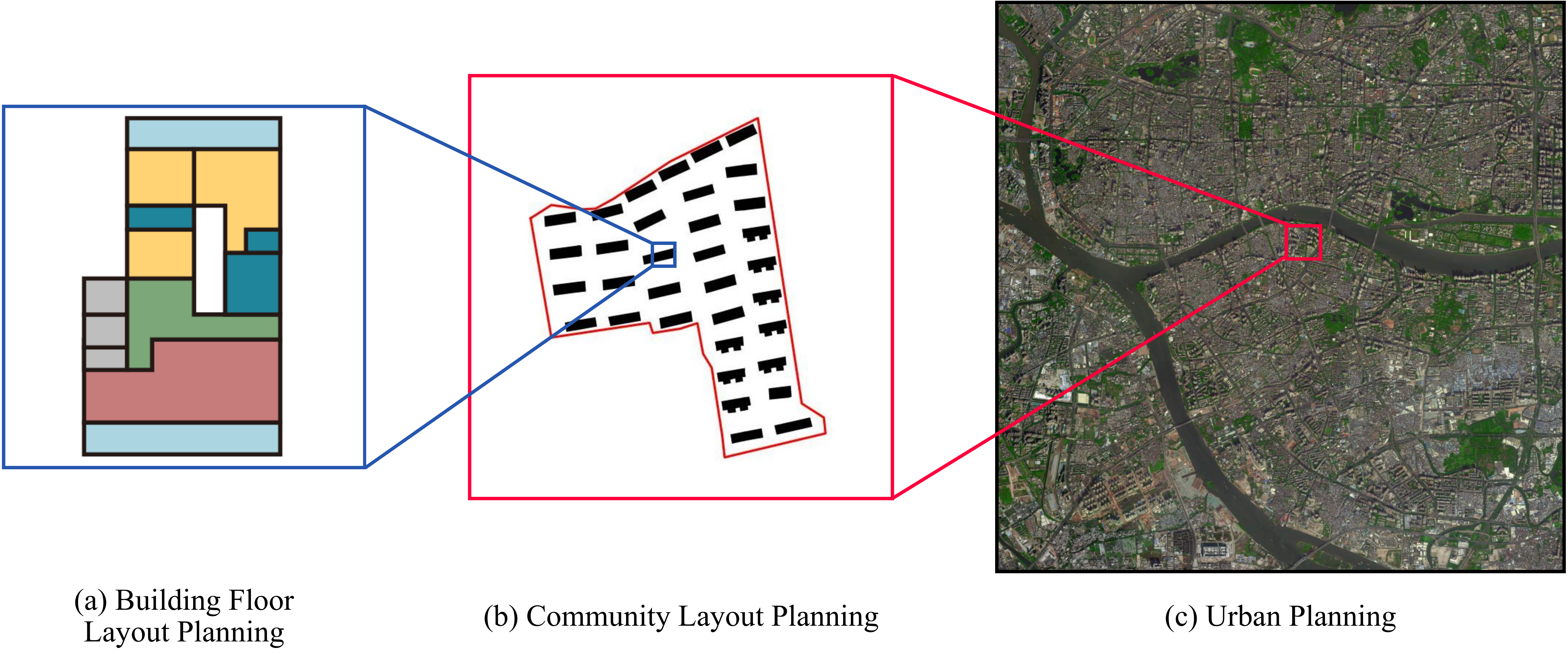}
\caption{Typical tasks of layout planning from fine- to coarse-grained (the projection relationship in the figure is for illustration only). (a) Building Floor Layout Planning \cite{nauata2020house}. (b) Community Layout Planning (an example generated from ReCo). (c) Urban Planning (an example screenshot from map.)}
\label{fig:plannings}
\end{teaserfigure}
%\protect\footnotemark


\maketitle

\section{Introduction}
\label{intro}
Layout tasks in architecture and urban planning refer to the physical arrangement of urban spatial components at different scales, in a creative and functionally reasonable way \cite{schneider2011rethinking}, where building floor layout planning, community layout planning and urban planning are the typical tasks from fine- to coarse-grained (shown in Fig. \ref{fig:plannings}), contributing significantly to the quality and sustainability of buildings, neighborhoods or entire cities. The essence of layout planning is an analytical and problem-solving activity that has to meet various requirements and specifications, while traditional expert-empirical-led planning methods are becoming sluggish in the increasingly complex contemporary design context. Recent years have witnessed a rapid growing research interest in intelligent design, e.g., design pattern recognition \cite{bei2019spatial, he2018recognition}, building volume generation \cite{osintseva2020automated}, and street network generation \cite{hartmann2017streetgan}, using advanced data-driven approaches, which greatly promote the quantitative digital layout planning tasks in a more automated, rational, and efficient way.

Among the three typical layout tasks, the research on building floor layout planning, represented by indoor furniture placement \cite{yu2011make, wang2018deep, wang2019planit, chaeibakhsh2021optimizing, paschalidou2021atiss} and floor plan generation \cite{as2018artificial, bao2013generating, nauata2020house, nauata2021house, sun2022wallplan}, is the most active, thanks in large part to the relatively high-quality mature datasets. In contrast, studies on data-driven urban planning and residential community layout planning are still limited \cite{jha2021review, chirkin2016concept}. As a key task that directly affects the quality of residence life and urban environmental space \cite{lai2014outdoor}, residential community layout planning plays a linking role between floor layout and urban planning. However, the effective work cannot be carried out widely due to the lack of large-scale, reliable, and open-source benchmark dataset. Specifically, current work on automatic generative residential layout and design mainly relies on rule-based approaches \cite{chirkin2016concept, miao2020development, osintseva2020automated, ying2021generating}. Although some efforts have been made to apply Generative Adversarial Networks (GAN) \cite{goodfellow2014generative} to generative design, the datasets involved cannot be accessed publicly \cite{sun2021}. Besides, the relevant analytic tasks, e.g., community layout pattern recognition \cite{bei2019spatial, he2018recognition, yan2019graph, zhang2014data, 8781683}, also leave the issue of limited performance and inadequate datasets. Despite some algorithms, e.g., online reinforcement learning \cite{sutton2018reinforcement}, show promising results to rely less on data or even require no historical data during training processes, sufficient data provided by dataset is still essential when it comes to model effectiveness evaluation \cite{alpaydin2016machine, li2017deep}. 

To resolve the data inadequacy issues and pave the way to robust and open data-driven modelling for residential community layout related tasks, in this paper, we introduce the \textbf{Re}sidential \textbf{Co}mmunity Layout Planning Dataset (\textbf{ReCo} Dataset), which is by far the first and largest vector dataset based on real-world residential communities. The ReCo Dataset contains 37,646 residential community layout plans sampled from 60 different cities covering 598,728 residential buildings. The height information of buildings is also included for the extension of two-dimensional (2D) information to 3D space, making ReCo applicable to more 3D modeling and spatial evaluation tasks \cite{8781683}. Unlike other raster image-based datasets in architectural design fields, e.g., LIFULL HOME's dataset \cite{lifull} and RPLAN dataset \cite{Wu_DeepLayout_2019}, ReCo provides more fine-grained coordinate information, through which commonly used raster data or vector (or polygon) data formats, such as image, Shapefile \cite{esri1998shapefile}, as well as 3D geometry (with height information) can be exported flexibly. By providing data in this form, the spatial attribute information of buildings, including distance and size, and the fundamental metrics of communities, including Floor Area Ratio (FAR) and Building Coverage Ratio (BCR) can be preserved and calculated, so that the dataset can be adapted to generation and analysis tasks at different granularity.

In addition to ReCo Dataset construction involving data collection and processing, this paper also demonstrates the dataset usability and benchmark in one of the principal downstream data-driven tasks, i.e., automated residential community layout planning, using the Deep Convolutional Generative Adversarial Network (DCGAN) \cite{radford2015unsupervised} and Conditional Generative Adversarial Nets (cGAN) \cite{mirza2014conditional} as backbone. The experiment results confirm the potential of applying our dataset for the tasks in architecture and urban design. 

The contributions of our paper can be summarized as follows:
\begin{itemize}
  \item We release ReCo, the \textbf{first large-scale open-source} residential community layout planning dataset. ReCo can be applied to numerous promising applications such as generative layout planning, pattern recognition, classification and spatial evaluation. 
  \item ReCo is a \textbf{diverse and extensive} dataset containing layout information of 37,646 residential communities and 598,728 buildings across 60 cities. These sample cities span a large geographical area covering inland cities, coastal cities, etc., with different urban characteristics.
  \item ReCo is a \textbf{fine-grained coordinate information-based} dataset that can be flexibly exported to various common spatial file formats. It provides more spatial attribute information than image-based datasets, so that can be applied to a wider range of tasks at different scales.
  \item We build two generative models to \textbf{validate dataset usability}, which can serve as baselines for benchmarking the task of automated residential community layout planning. We believe ReCo has a great potential to expedite research in the growing field of intelligent layout planning.
\end{itemize}

\section{Related work}
In this section, we firstly review the related methods of the three typical layout tasks, namely community layout planning, building floor layout planning and urban planning. Then we conclude the datasets applied to these tasks. In general, the maturity and scale of existing datasets vary a lot in terms of layout types, resulting in inconsistent development of technical methods in the field. It is particularly time-consuming and challenging to collect high-quality data for community layout research, thus there are only a few small-scale datasets at community level.

\paragraph{Methods.} The study of building floor layout planning has been at the forefront of the three, with numerous efforts using advanced algorithms to artificial intelligence task, which has largely replaced the traditional experience-oriented design method. Earlier research is mainly based on computer-aided methods by exploiting explicit rules, i.e., translating domain prior knowledge into computer algorithms \cite{egor2020computer, ma2014game, merrell2010computer, chaeibakhsh2021optimizing}.
However, methods based on finite rules are bound to be difficult to deal with the complex relational modeling in layout planning, the development of related datasets and models provides solutions to this challenge \cite{bengio2009learning, anthes2013deep}. 
%%%%%%%%%%%%%%%%%%%%%%%%%%%%%%%%%%%%%%%%%%%%%%%%%%%%%%%%%%%%%%%%%%%%%%%%%%%%%%%%
As for community layout planning tasks, the existing research mainly focuses on community layout pattern recognition and classification \cite{8781683, bei2019spatial}. In contrast with building floor layout planning, there are few studies on community generative design. The development of this field is hindered by the small-scale, closed-source or proprietary datasets \cite{osintseva2020automated, sun2021, li2019layoutgan}.
%%%%%%%%%%%%%%%%%%%%%%%%%%%%%%%%%%%%%%%%%%%%%%%%%%%%%%%%%%%%%%%%%%%%%%%%%%%%%%%%
For the coarse-grained tasks of urban planning, most of the current research focuses on design optimization, generative design, and urban environmental evaluation \cite{miao2020development, hartmann2017streetgan, liu2017machine, chirkin2016concept}. However, commonly used machine learning methods in the field of Computer Vision (CV), e.g., Variational Auto-Encoder \cite{doersch2016tutorial} and GAN are under-utilized in design optimization \cite{miao2020development}. Only a few methods are developed to generate road networks, but exhibit potential to enhance generative urban design \cite{hartmann2017streetgan}. 
%Similar approaches may be applicable to the planning of urban blocks, plots and buildings. 
As for urban environmental evaluation, when numerous objective evaluation metrics are required, only a concept of interactive machine learning integrating clustering, feature extraction, and human subjective-oriented Reinforcement Learning \cite{sutton2018reinforcement} has been proposed \cite{chirkin2016concept} where adequate data is particularly important to help establish the objective reward function and evaluation system.

\paragraph{Datasets. \label{related_datasets}} Two commonly used open-source datasets for building floor layout planning tasks are \textit{RPLAN dataset} \cite{Wu_DeepLayout_2019} and \textit{LIFULL HOME dataset} \cite{lifull}, which offer 80K annotated house layouts and 5 million ground-truth floorplans, respectively. They have been applied to automatic floorplan generative models \cite{hu2020graph2plan, nauata2020house, nauata2021house, wang2021actfloor, sun2022wallplan}. With large-scale datasets provide the sufficient training data for GAN, existing models can automatically formulate floor plans that are indistinguishable from the ground-truth \cite{nauata2020house, nauata2021house}.  
%%%%%%%%%%%%%%%%%%%%%%%%%%%%%%%%%%%%%%%%%%%%%%%%%%%%%%%%%%%%%%%%%%%%%%%%%%%%%%%%
Community layout planning models are struggling with limited amount of data. For example, \textit{Dong et al.} \cite{8781683} proposed a Convolutional Auto-Encoder model to embedding plots by applying a dataset with 1,887 samples. The study suggests that larger-scale data covering more attributes can help improve model performance and utility. \textit{Bei et al.} \cite{bei2019spatial} introduced Graph Convolutional Network (GCN) \cite{kipf2016semi} to accomplish different tasks of building state identification, building node clustering and building pattern recognition. Nevertheless, the individual blocks containing building contours in dataset are randomly selected rectangular areas rather than strict community boundaries. In addition, \textit{Yan et al.} \cite{yan2019graph} presented a GCN to classify building patterns which also remains limited by the small dataset (2,194 samples). For the work on intelligent community layout planning, \textit{Cheng et al.} \cite{sun2021} applied the Convolutional GAN to generate residential layout planning while the training process is still limited by the small sample size of 1,050.
The diversity of data is also crucial for pattern recognition and layout generation tasks since it can help provide more sufficient information and wider range of data generation distribution \cite{wang2009diversity, creswell2018generative}. Furthermore, due to the lack of benchmark datasets, it is difficult to evaluate and compare the performance of different models.
In summary, data-driven community layout planning tasks rely heavily on datasets. 
%%%%%%%%%%%%%%%%%%%%%%%%%%%%%%%%%%%%%%%%%%%%%%%%%%%%%%%%%%%%%%%%%%%%%%%%%%%%%%%%
In terms of urban planning, Hartmann et al. \cite{hartmann2017streetgan} applied data extracted from \textit{OpenStreetMap (OSM)} \footnote{\url{https://www.openstreetmap.org/} \label{osm}} to generate road networks. 
Although \textit{OSM} contains a large amount of raw geographic data, it cannot be directly applied to model training without complex preprocessing.
%%%%%%%%%%%%%%%%%%%%%%%%%%%%%%%%%%%%%%%%%%%%%%%%%%%%%%%%%%%%%%%%%%%%%%%%%%%%%%%%
We summarize and compare the above datasets and ReCo dataset, as shown in Tab. \ref{tab:datasets}.

\paragraph{Comparing three types of layout planning.} Generally, the smaller scale building floor layout planning has been more widely studied, especially %with the combination of GANs 
in generation tasks, benefiting from numerous relatively mature datasets; while the development of the larger-scale of urban planning and community layout planning is subject to the lack of high-quality benchmark datasets. %Specifically, the urban planning tasks are mostly implemented semi-automatically, involving rules and human intervention for the establishment of relevant constraints or evaluation metrics. 
Regarding community layout planning, there is still a lot of room for improvement in the performance of data-driven models, which highlights the urgent need of large-scale open-source benchmark datasets for this research area. We hope to tackle the field development issues through the release of ReCo Dataset, thereby accelerating the maturity of data-driven methods for community layout planning, and even for the smaller- or larger-scale layout tasks in the field of architecture design. 
%%%%%%%%%%%%%%%%%%%%%%%%%%%%%%%%%%%%%%%%%%%%%%%%%%%%%%%%%%%%%%%%%%%%%%%%%%%%%%%%

\section{ReCo Dataset}
\label{dataset}
In this section, we describe the unique characteristics of ReCo, as well as the pipeline to acquire the dataset from the collection of raw data in real-world residential communities, to the calibration and standardization of community boundaries and building outlines. 

\subsection{Properties of ReCo Dataset}
The ReCo dataset is based on high-precision vector coordinates rather than raster images, which can be easily converted to various data types, such as 2D image, 3D geometry and Shapefile. The properties of ReCo can be summarized in the following three major points:

%\paragraph{Diversity.} ReCo covers residential community layout data in 60 cities of different locations and sizes. See Section \ref{Dataset Stats} for specific statistics on ReCo. The diversity of the ReCo allows researchers to categorize datasets based on features such as location, number of buildings, and plot size for different research purposes.

\paragraph{Diversity.} ReCo covers residential community layout data in 60 cities, with different scales of residential areas, residential distribution characteristics, and residential forms, which constitute the diversity of the dataset. See Section \ref{Dataset Stats} for specific statistics. ReCo allows researchers to classify datasets for different research purposes based on features that are not limited to location, number of buildings, and plot size.

\paragraph{Flexibility.} To our knowledge, ReCo is so far the first and largest open-source vector dataset related to real-world community. The form of the dataset based on coordinates makes it flexible to output various common spatial data formats, and retain the original information of data to the greatest extent.

\paragraph{Uniformity.} ReCo can serve as a standard dataset for the residential community layout planning related tasks, providing a benchmark for evaluating the performance of various data-driven models, to facilitate the convergence and progress of advanced algorithms and urban planning.

\subsection{Data format and description}

\begin{figure}[!htbp]
\centering
\includegraphics[width=1\linewidth]{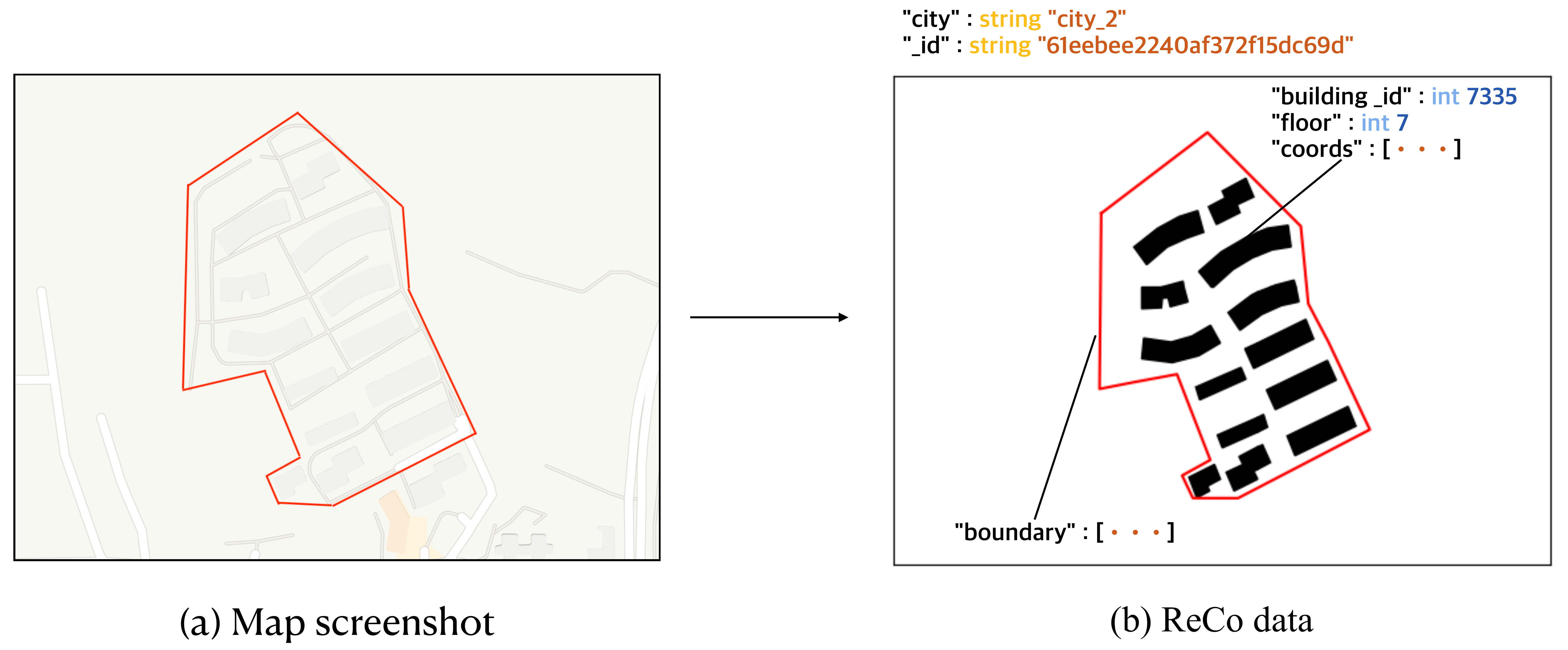}
\caption{Examples of image data. (a) an example community screenshot in a city from map. (b) the community from ReCo Dataset corresponding to (a).}
\label{fig:data_example}
\end{figure}

% \textsuperscript{\ref{applemaplink}}

% \footnotetext{Screenshot from Apple Map (\url{https://www.apple.com/maps/}), while city maps are considered as raw dataset.\label{applemaplink}}

ReCo is provided with data-interchange formats of JSON and GeoJSON \cite{butler2016geojson} that describes the information of coordinates and spatial attributes. These types of vector data formats can support commonly used spatial format conversions. To make it easier for users to apply ReCo to image-based layout tasks, we provide a way to generate the 2D image from existing dataset formats, with the code published at GitHub\footnote{Related code is at our GitHub repository: \url{https://github.com/FDUDSDE/ReCo-Dataset.} \label{github}}.

To explain the content of the ReCo Dataset in detail, we randomly select an example community (shown in Fig. \ref{fig:data_example} (a)) and convert the corresponding data in ReCo Dataset into 2D image (shown in Fig. \ref{fig:data_example} (b)).
ReCo consists of three types of instances, namely building, residential community and city. While the basic elements for describing the instances are coordinates. We summarize the basic element and instance types as following (a more detailed example of data instances is shown in Appendix. \ref{app_instance}):
\begin{itemize}
\item Coordinate: geographical coordinates have been converted to Mercator coordinates \cite{maling2013coordinate} and desensitized for legal and privacy concerns.
\item Building: residential buildings arranged in the community which consists of a set of coordinates describing the outline, and each building has a unique identifier (``building\_id'') within its city limits. The building height attribute is represented by ``building floor'', with an assumption that each floor is 3 meters high.
\item Community: the community where buildings are located which can be recognized by the ``\_id'' (the unique identifier, automatically generated by MongoDB\footnote{\url{https://www.mongodb.com}}) with city attribute to explain the location. A set of coordinates (community boundary coordinates) describe community's outline, and the value of coordinates is constrained to be non-negative.
\item City: the city where communities are located. In the ReCo, ``city'' is also considered as one of the attributes of communities. A set of community instances with the same ``city'' attribute is a sub-set of the dataset.
%sub-set of the dataset, and city information is used as label for each community. A set of community instances with the same ``city'' label can be considered as a special instance. % A set of communities with the same ``city'' label are sampled from the same city in real world.
\end{itemize}

\subsection{Data collection and generating pipeline}
\label{collection}
\begin{figure*}[!htbp]
\centering
\includegraphics[width=1\linewidth]{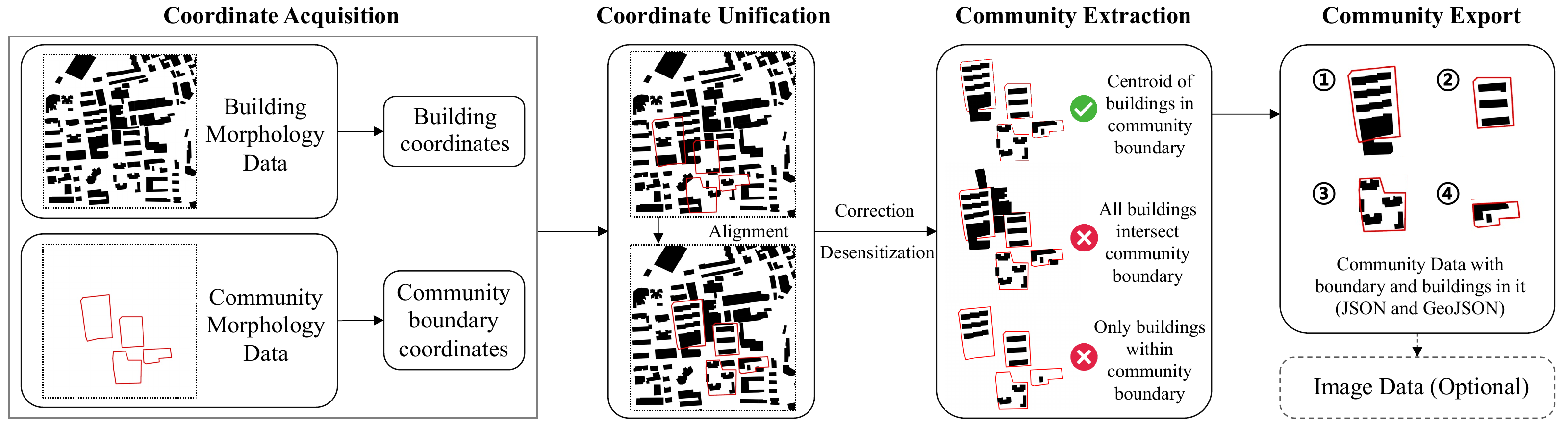}
\caption{Dataset generating pipeline.}
\label{fig:flow}
\end{figure*}
In order to capture the morphology of residential community layout plan on a large scale, we prepare two parts of raw data. The first is the information of buildings in the map including buildings' vector outline and height information (as Building Morphology Data) collected from OpenStreetMap \textsuperscript{\ref{osm}}  and Google Earth Engine\footnote{\url{https://developers.google.com/earth-engine} \label{google}}. The second is the residential community information including boundary coordinates information (as Community Morphology Data) acquired through the Baidu Map APIs\footnote{\url{https://lbsyun.baidu.com} \label{baidu}}.

The visualized dataset generation pipeline is presented in Fig. \ref{fig:flow}. As shown in Fig. \ref{fig:flow}, after two parts of raw data are collected, the information of corresponding coordinates is extracted. Since the geographic coordinate systems of these multi-source data are different, unification is required to align the two parts of coordinate data. The data from different geographic coordinate systems are projected onto the same 2D plane, i.e., the Transverse Mercator projection \cite{maling2013coordinate}. Due to the sensitivity of the Geo information, coordinate correction and desensitization are added after the unification process. Next, we are allowed to determine whether the building belongs to the community by calculating whether the building centroid is within the area enclosed by the community boundary, under the unified spatial environment. Finally, the two parts of data are combined into one as the pipeline output, which completely describes the information of residential community layout plannings.
% The visualized dataset generation pipeline is presented in Fig. \ref{fig:flow}. As the Fig. \ref{fig:flow} shows, after two parts of raw data are collected, they are combined into one, which completely describes the information of urban residential communities. Due to the different geographic coordinate systems of these multi-source data and the sensitivity of the Geo information, coordinate correction, unification and desensitization are required in the combination process, which is also the main task of data cleaning. The data from different geographic coordinate systems are projected onto the same 2D plane, i.e., the Transverse Mercator projection \cite{maling2013coordinate}. Next, we are allowed to determine whether the building belongs to the community by calculating whether the building centroid is within the area enclosed by the community boundary, under the unified spatial environment. 
%%%%%%%%%%%%%%%%%%%%%%%%%%%%%%%%%%%%%%%%%%%%%%%%%%%%%%%%%%%%%%%%%%%%%%%%%%%%%%%%
In addition, the information of the building height is also kept. We save the data in JSON and GeoJSON formats for users to export images or Geographic maps. 
%%%%%%%
% Since most models applied in the community planning field are based on image format data, we add an extra optional step, i.e., a data visualization method to convert the raw data into 2D images, at the end of the data generation process. 
%%%%%%%
% Example of visualized data in RGB image is shown in Fig. \ref{fig:data_example} (b), and the relevant scripts are published on GitHub \textsuperscript{\ref{github}}. 

\subsection{Dataset statistics}
\label{Dataset Stats}
\begin{table}[!htbp]
%\small
\centering
\caption{Statistics of ReCo Dataset.}
\begin{tabular}{cccc}
  \toprule
  Stats & \# of Communities      & \# of Buildings        & Average         \\ 
  \midrule
  Max   & 3,947 & 70,614 & 23.40 \\
  Min   & 7    & 143   & 10.35 \\
  Mean  & 627.43          & 9,978.80         & /               \\
  Total & 37,646          & 598,728          & 15.90           \\ 
  \bottomrule
\end{tabular}
\label{tab:stats}
\end{table}

\begin{table*}[!htbp]
%\small
\centering
\caption{Datasets comparison. The ReCo Dataset is provided in JSON and GeoJSON formats originally, with support of conversions to commonly used spatial format.}
\begin{threeparttable}[b]
    \begin{tabular}{cccccc}
      \toprule
      Tasks                          & Dataset/ Paper                                    & Data type & \# of samples & Sampled from & Accessibility \\ \midrule
      \multirow{2}{*}{\begin{tabular}[c]{@{}c@{}}Building floor\\ layout planning\end{tabular}} & RPLAN \cite{Wu_DeepLayout_2019}                & Images    & 80K           & /            & Open-source  \\
                              & LIFULL\cite{lifull}                                 & Images    & 5M            & /            & Open-source  \\ \midrule
      \multirow{5}{*}{\begin{tabular}[c]{@{}c@{}}Community \\ layout planning\end{tabular}}      & Dong et al. \cite{8781683}       & Images    & 1,887         & 1 city       & Private      \\
                              & Bei et al. \cite{bei2019spatial} & Vectors    & 1,304+        & 2 cities     & Private      \\
                              & Yan et al. \cite{yan2019graph}   & Images    & 2,194         & 2 cities     & Private      \\
                              & Cheng et al. \cite{sun2021}      & Shapefile    & 1,050         & /  & Private      \\
                              & \textbf{ReCo (Ours)}                                       & JSON      & 37,646        & 60 cities    & Open-source  \\ \midrule
      Urban planning                         & OSM \textsuperscript{\ref{osm}}                                     &  *   & /        & /   & Open-source  \\ \bottomrule
  \end{tabular}
  \begin{tablenotes}
    \item[*] OSM is considered as raw data and there is no processed public datasets for urban planning tasks.
  \end{tablenotes}
\end{threeparttable}
\label{tab:datasets}
\end{table*}

In order to demonstrate the ReCo more specifically, we provide statistics of the ReCo (shown in Tab. \ref{tab:stats}). 
%%%%%%%
% ReCo contains a total of 37,646 residential community layout plans from 60 cities in China, including 598,728 marked residential buildings, with an average of 15.90 buildings in each residential community. 
%%%%%%%
Significantly, all data in ReCo comes from the real world and meets the realistic requirements for the layout of residential buildings.
%%%%%%%%%%%%%%%%%%%%%%%%%%%%
Comparing the datasets for the three typical tasks mentioned in Section \ref{related_datasets}, the ReCo dataset is by far the largest and the only open-source dataset in the community layout planning field (shown in Tab. \ref{tab:datasets}). 
Moreover, the data volume of ReCo has increased by more than 17 times compared to previous largest residential layout dataset \cite{yan2019graph}. In addition, ReCo has the widest data distribution with samples from 60 different cities, which increases the diversity of the dataset. However, compared with the two datasets \cite{Wu_DeepLayout_2019, lifull} commonly used in the building floor layout planning, ReCo still has a huge room for improvement in data volume. % We also hope that ReCo will inspire more work on dataset expansion.
% Moreover, in this field, ReCo has increased the data size by more than 17 times than the previous largest dataset which applied by Yan et al. \cite{yan2019graph}. In terms of data distribution, ReCo also samples from the widest range of 60 different cities increasing the diversity of data samples.
% However, compared with the two commonly used open-source datasets in the field of building floor layout planning, ReCo still has a huge room for improvement in data volume. We also hope that the ReCo can inspire more work on dataset expansion.

%For the statistics of each city, we use random number as the city name labels considering the desensitization issue. The city with the largest amount of data samples is ``city\_60'', which has 3,947 data samples, including 70,614 buildings, with an average of 17.89 buildings in each residential community. The ``city\_60'' will be used for the following experiments.

\section{Experiments}
\label{experiments}

The GAN \cite{goodfellow2014generative} models have achieved a breakthrough in the field of building floor layout planning\cite{as2018artificial, nauata2020house, nauata2021house}. We expect that GANs can also be applied to residential community layout planning generation tasks if supported by sufficient data. Based on the ReCo Dataset, therefore, we propose a residential community layout planning generative model, and conduct a baseline experiment. 
%%%%%%%%%%%%
In addition, the residential community layout planning tasks are often subject to various constraints, e.g., community boundary constraints, in the actual design process. Therefore, we propose another generative model for residential community layout planning based on boundary constraints.
%%%%%%%%%%%%
We aim to answer the following research questions:
\begin{itemize}
\item \textbf{RQ1:} Can our dataset be applied to residential community layout planning generative tasks, and how does it perform?
\item \textbf{RQ2:} How does the size of training dataset affect the performance of residential community layout planning generative model?
\item \textbf{RQ3:} What is the effect of different data distribution (i.e., sampled from different regions) on training of the model?
\item \textbf{RQ4:} Can our dataset be applied to generative tasks based on community boundary constraints, and how does it perform?
\item \textbf{RQ5:} Which model performs better? The model with or without the boundary as constraint?
\end{itemize}

\subsection{Residential community layout planning generative model (RQ1,2,3)}

\begin{figure}[!htbp]
\centering
\includegraphics[width=1\linewidth]{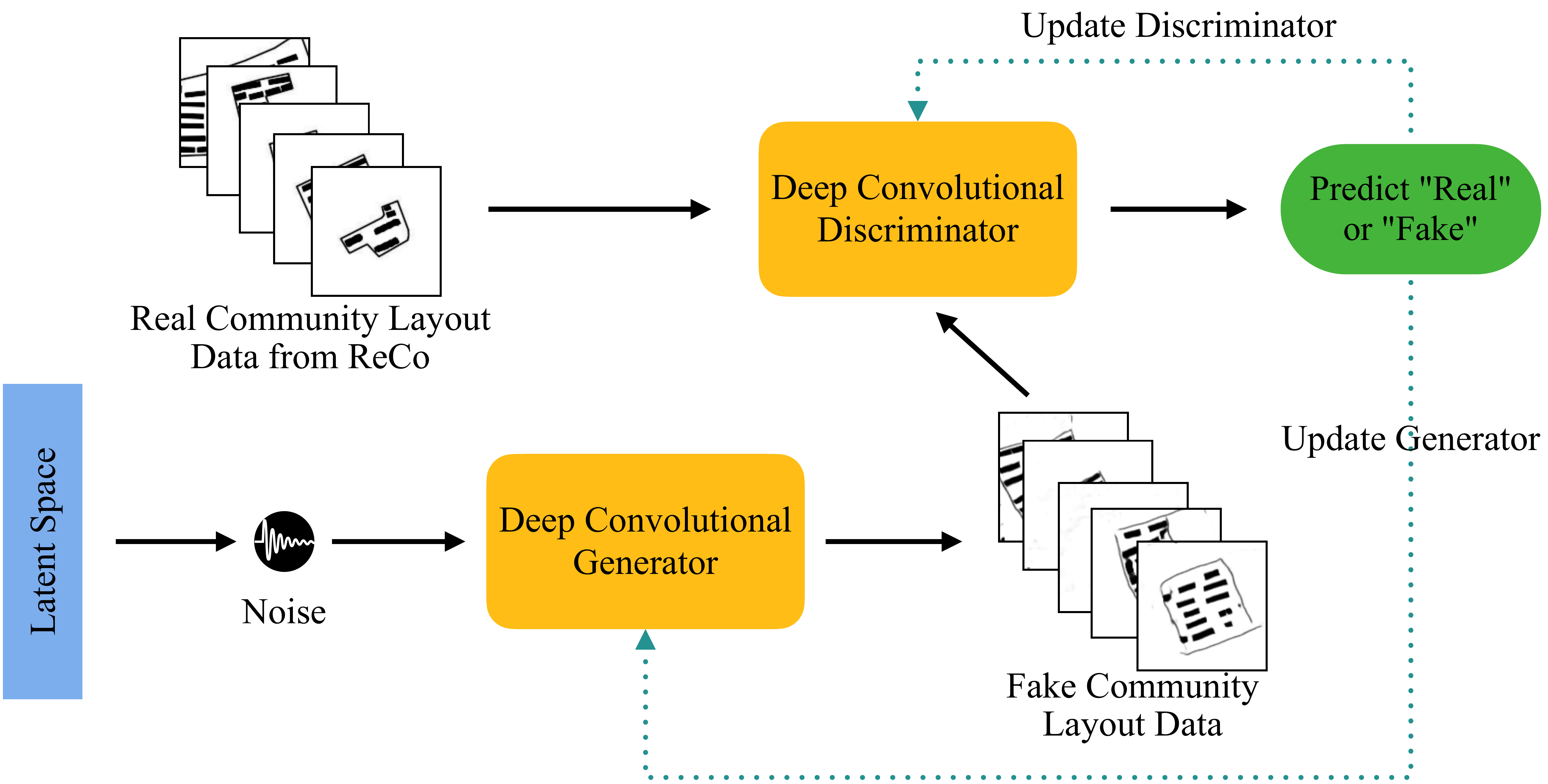}
\caption{Residential community layout planning generative model architecture.}
\label{fig:model}
\end{figure}

\begin{table}[!htbp]
\centering
\caption{The datasets used in our experiments.}
\begin{tabular}{ccc}
\toprule
Dataset        & Description      & Size                       \\ \midrule
city\_60   & the data from the $60^{th}$ city in the ReCo  & 3,974 \\
h\_city\_60  & randomly sampled from the city\_60 & 2,000 \\
city\_40      & the data from the $40^{th}$  city in the ReCo     & 2,095      \\
ReCo   & the whole ReCo Dataset   & 37,646  \\
\bottomrule                            
\end{tabular}
\label{tab:subsets}
\end{table}

We trained a DCGAN-based \cite{radford2015unsupervised} generative model for residential community layout planning by applying ReCo, to demonstrate the applicability of our dataset (our experimental code is redeveloped based on the GitHub repository\footnote{\url{https://github.com/eriklindernoren/PyTorch-GAN}}). The model architecture is illustrated in Fig. \ref{fig:model}. ReCo and its three subsets used in the experiments are summarized in Tab. \ref{tab:subsets}. The model was trained for 2K epochs with a batch size of 128 per sub-experiment. % Detailed training information can be found in our repository\textsuperscript{\ref{github}}.
% More detailed pre-processing and training information can be found in our GitHub repository\textsuperscript{\ref{github}}.

\subsection{Boundary constrained residential community layout planning generative model (RQ4,5)}
\begin{figure}[!htbp]
\centering
\includegraphics[width=1\linewidth]{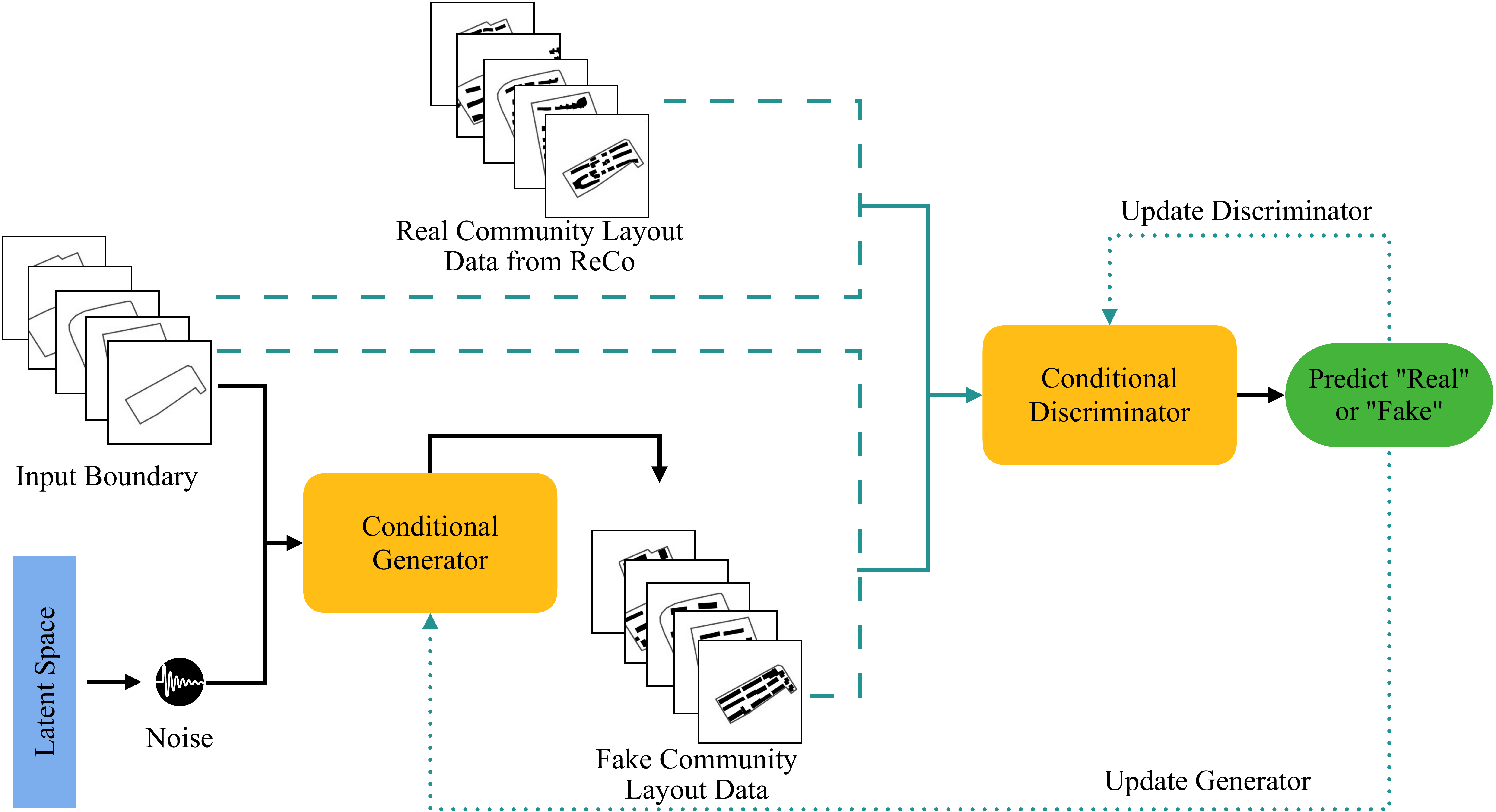}
\caption{Boundary constrained residential community layout planning generative model architecture.}
\label{fig:cmodel}
\end{figure}

To further meet the design requirements, we took community boundaries as input constraints and trained a cGAN- and pix2pix-based \cite{mirza2014conditional, isola2017image} conditional residential community layout planning generative model by applying ReCo (The code is redeveloped based on the GitHub repository\footnote{\url{https://github.com/junyanz/pytorch-CycleGAN-and-pix2pix}}). The model architecture is shown in Fig. \ref{fig:cmodel}. We used the same datasets in Tab. \ref{tab:subsets}. The model was trained for 2K epochs per sub-experiment. % Detailed training information can be found in our repository\textsuperscript{\ref{github}}.

\subsection{Results}
% To illustrate the results of the model in an intuitive perspective, we firstly sample the generated images (of the model trained with the complete ReCo) and present them in Fig. \ref{fig:pre-processed&generated}. By comparing with the ground-truth data, we can conclude that part of the generated data has the morphological characteristics of the real data. However, existing models do not perform very well in the case of uneven building spacing and irregular boundaries. There still remains challenge for researchers to build models to optimize for this scenario, and the ReCo provides the data basis for this work.

\paragraph{Model evaluation} To evaluate the performance of the GAN model, it is common practice to use the Fréchet Inception Distance (FID) scores \cite{heusel2017gans, Seitzer2020FID}, which measures the distance between the distribution of real data and generated data (a smaller FID score means the generated data is closer to the real data). This evaluation method has also been used in HouseGAN++ \cite{nauata2021house}, which is one of the typical models in the field of building floor layout planning. % We compare the four experiments by dividing them into two subgroups and the experiment results measured by FID scores are shown in Fig. \ref{fig:results} (a) and (b) respectively. 

\begin{figure}[!htbp]
\centering
\includegraphics[width=1\linewidth]{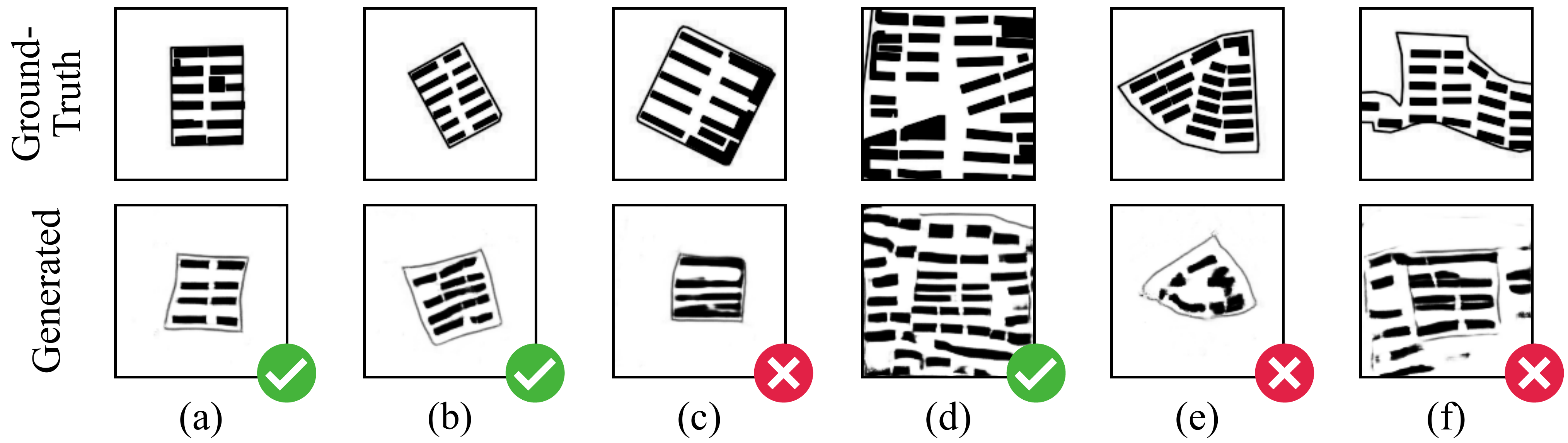}
\caption{Examples of ground-truth images and generated images of community layouts. A green mark denotes the preferred design.}
\label{fig:pre-processed&generated}
\end{figure}

\paragraph{Generative performance of the model without boundary constraint (RQ1)} To demonstrate the model generation performance based on ReCo, we sample the images generated by model trained on the complete ReCo (shown in Fig. \ref{fig:pre-processed&generated}). As shown in Fig. \ref{fig:pre-processed&generated} (a), (b) and (d), we can conclude that part of the generated data has the morphological characteristics of the real data. However, as shown in Fig. \ref{fig:pre-processed&generated} (c), there still are communities with uneven building spacing in the generated results. Also, Fig. \ref{fig:pre-processed&generated} (e) and (f) show that the existing model performs poorly in the generation of communities with irregular boundaries. We build the boundary constrained model to optimize this situation. % Optimizing for this situation remains a challenge, and ReCo provides the data foundation for it. 

% By comparing with the ground-truth, we can conclude that part of the generated data has the morphological characteristics of the real data (as shown in Fig. \ref{fig:pre-processed&generated} (a), (b), (d)). However, the existing model performs poorly in the generation of communities with uneven building spacing (as shown in Fig. \ref{fig:pre-processed&generated} (c)) and irregular boundaries (as shown in Fig. \ref{fig:pre-processed&generated} (e), (f)). Optimizing for this situation remains a challenge, and ReCo provides the data foundation for it. 

\begin{figure*}[!htbp]
\centering
\includegraphics[width=1\linewidth]{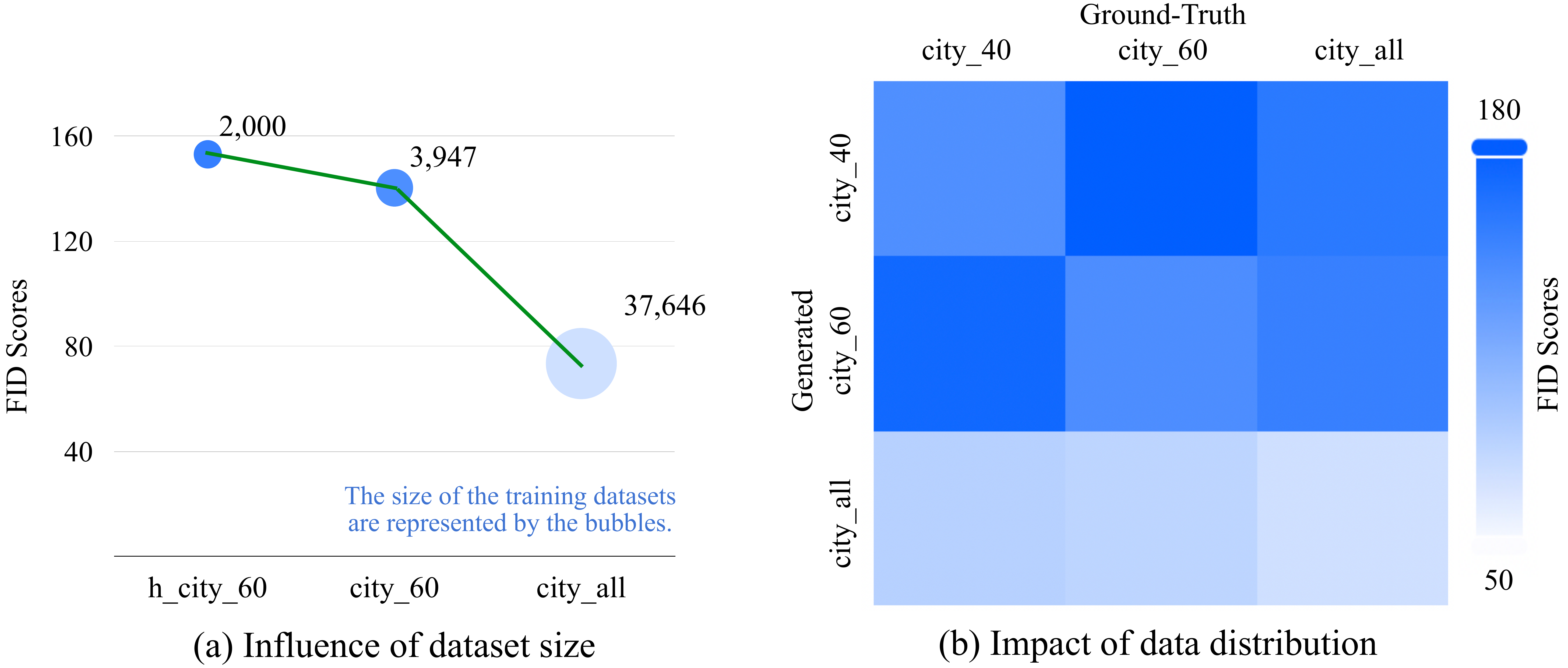}
\caption{Influence analysis. (FID Scores, the-lower-the-better.)} %We use the trained models to generate 5,000 ``fake'' images respectively, and cross-compare with the training data. (b) should be read row-by-row.
\label{fig:results}
\end{figure*}

\paragraph{Influence of dataset size (RQ2)} As shown in Fig. \ref{fig:results} (a), we observe that the FID score decreases, i.e., the performance of model increases, as the dataset size increases. Besides, it can be found in Fig. \ref{fig:results} (b), that the model trained on the ReCo has better performance than trained only on ``city\_60''. Similarly, the performance of the model trained on ``city\_60'' is better than trained on ``city\_40''. These observations indicate that the effect of the dataset size on the experiment, that is, in the case of the same data distribution, the more training data, the better the experimental results.

%By comparing the experiment results of models trained on ``h\_city\_60'', ``city\_60'' and ReCo (shown in Fig. \ref{fig:results} (a)), we can draw the effect of the dataset size on the experiment, that is, in the case of the same data distribution, the more training data, the better the experimental results. This observation can also be found in Fig. \ref{fig:results} (b), where the model trained on full amount of data (``city\_all'') has better performance than trained on ``city\_60''. Similarly, the performance of the model trained on ``city\_60'' is better than trained on ``city\_40''.

% By comparing the experiment results of models trained with ``h\_city\_60'', ``city\_60'' and ``city\_all'', where the data are sampled from the same parent dataset (ReCo) (Fig. \ref{fig:results} (a)). We can see that as the amount of training data increases, the model performs better (i.e. lower FID scores). This reflects the impact of the dataset size on the experiment, i.e., with the same data distribution, the more training data brings better experimental results. This observation can also be found in Fig. \ref{fig:results} (b), where the model trained with full amount of data (``city\_all'') has better performance than trained with ``city\_60''. Also, the performance of the model trained with ``city\_60'' is better than trained with ``city\_40''.

\paragraph{Impact of data distribution (RQ3)} As shown in Fig. \ref{fig:results} (b), for the diagonal grid, the score is the lowest in each row. This demonstrates that the FID score between generated data and the corresponding ground-truth data is the lowest.
From the perspective of columns, we can see that the lowest scores appear all in the last row, which means the model trained on sufficient data (ReCo) performs better than the model trained on single city data. Moreover, the high scores are seen when comparing data from different cities, e.g., comparing the generated data of model trained by ``city\_40'' to ground-truth data of ``city\_60''. From these observations, we can conclude that the generated data distributions are closer to the corresponding ground-truth data and there is a certain gap in the data distribution of different cities. This also reflects the diversity of our datasets. However, in the case of sufficient training data, the influence of different training data distribution can be gradually ignored. 

% The observation of high scores between data from different cities demonstrates that there is a certain gap in the data distribution of different cities, which reflects the diversity of our datasets.
% From these observations, we can conclude that the generated data distributions are closer to the corresponding ground-truth data. However, in the case of sufficient training data, the influence of different training data distribution can be gradually ignored. 

% The result in Fig. \ref{fig:results} (b) shows the generated data distributions are all closer to the original corresponding ground-truth data. Moreover, comparing the data generated by different training data with the other two sets of ground-truth data, we can see the difference of FID scores. This demonstrates that for different cities, their data distribution also has a certain gap, which also reflects the diversity of our dataset. However, the model trained by enough data (``city\_all'') performs better than the model trained by single cities' data. This demonstrates that the distributions of different city are distinct, however, with enough training data, the impact on the training results due to different data distribution can be gradually omitted.

\begin{figure*}[!htbp]
\centering
\includegraphics[width=1\linewidth]{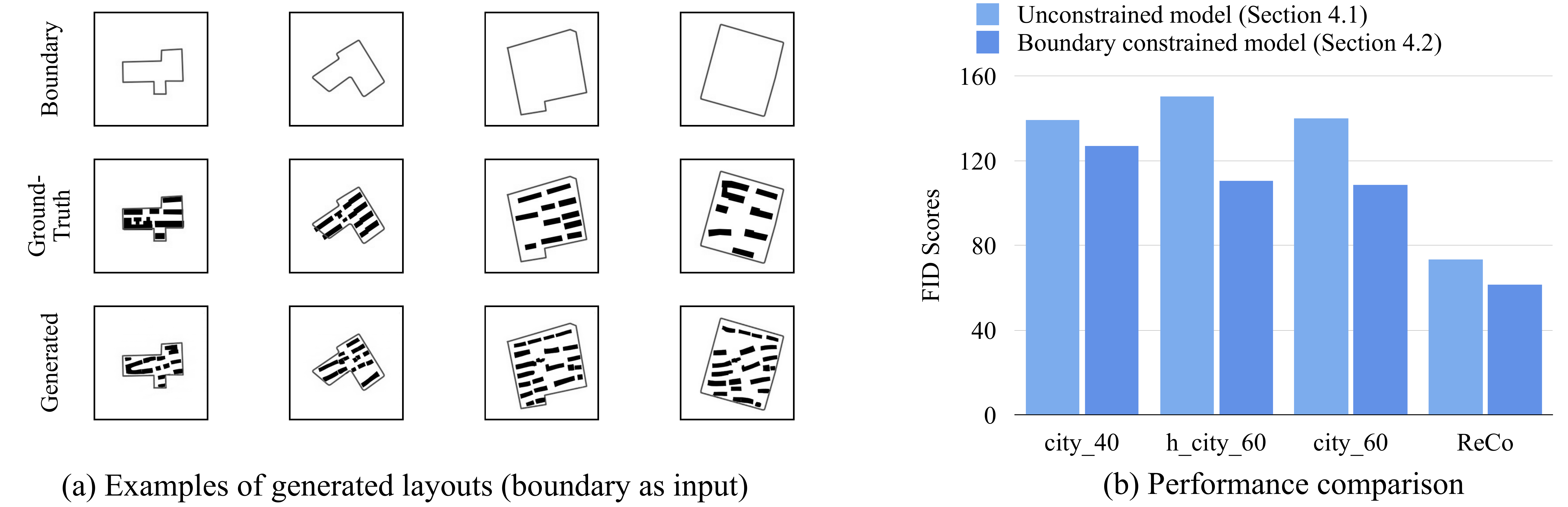}
\caption{(a) Results of boundary constrained model and (b) model comparison.}
\label{fig:results2}
\end{figure*}

\paragraph{Performance of boundary constrained model. (RQ4 and RQ5)} We sample images generated by the boundary-constrained model trained on the complete ReCo which is demonstrated in Fig. \ref{fig:results2} (a). By comparison with the results in Fig. \ref{fig:data_example}, it can be concluded that the model is able to correctly identify the boundaries of the community and place buildings within the boundary. Furthermore, it performs better in generating communities with irregular boundaries. From the FID scores shown in Fig. \ref{fig:results2} (b), we can also conclude that the boundary-constrained model performs better on all four datasets. In practice, community layout planning is also constrained by other indicators, such as FAR, BCR, etc. We speculate that the performance of the model can be further improved in the modeling case with more constraints.

% We sample the images generated by the boundary constrained model trained on the complete ReCo and demonstrated in Fig. \ref{fig:results2} (a). By comparing with the results in Fig. \ref{fig:data_example}, we can see that the model has been able to correctly identify the boundary of communities and place the building within the boundary. Also, it has a better performance on the generation of communities with irregular boundary. By observing the FID scores shown in Fig. \ref{fig:results2} (b), we can conclude that the boundary constrained model performs better on all four datasets. In practical situations, the community layout planning is constrained by other metrics, e.g. FAR, BCR etc. We speculate that in the case of modeling with more constraints, the generation performance can be further improved.

\section{Discussion}
%In this section, we discuss the limitations of ReCo dataset and future work that can be carried out. % based on it. 
\paragraph{Limitations.\label{limitation}} We provide a desensitization dataset ReCo based on coordinate and spatial attribute information, which can be flexibly exported to multiple common spatial file formats. The potential of the ReCo to help researchers build relevant models for the residential community task is also demonstrated in the study. However, there is still some room for improvement in this work. For instance, the Points of Interest (POI) around the community can help researchers obtain contextual information such as functionality, but such information has not been included in the dataset. The height information in our dataset is based on an assumption of average floor height rather than precise building height information. More precise building height information can help researchers to conduct more sophisticated studies.
% Since the height information in our dataset is based on an assumption of average floor height, precise building height information has not been described. More precise building height information can help researchers to conduct more sophisticated studies.
This dataset is somewhat of a mapping of the real-world communities, which means that the dataset should be updated over time. However, most studies are based on historical data, which does not affect research that applies our current dataset. Nonetheless, constantly updating, improving, and maintaining datasets remains a challenge.

\paragraph{Future work.\label{Future Work}} The ReCo Dataset can be extended in the future by collecting and adding more raw data, and can be classified into sub-datasets according to different attributes, such as geographical environment, and community area. % and building number. 
Experimental results show that there still are plenty of room to improve the planning effects. This indicates challenges remained in applying ReCo and better models with specific designs are welcome. 
Our dataset currently only covers residential buildings, we would like to expand the dataset by including other building types, e.g., commercial buildings, and urban complexes, to stimulate more related work, including urban design with different scales and mixed functions. 
Furthermore, ReCo can also be applied to multiple architectural design tasks, such as obtaining evaluation metrics for designs, and evaluating performance of design schemes or models.

\section{Conclusion}
In this paper, we introduce \textbf{Re}sidential \textbf{Co}mmunity Layout Planning (\textbf{ReCo}) Dataset, a novel scalable open-source vector dataset related to real-world communities. The current version of the dataset contains 37,646 community layout plans across 60 cities, covering 598,728 residential buildings. The building height information is also included for the extension of 2D information to 3D space.
Moreover, we demonstrate the great potential of data-driven models for the automatic generation of community layouts based on our dataset. 
% However, work on expanding and refining the dataset and developing more promising models remains a challenge. 
We expect that our dataset will stimulate extensive research on data-driven approaches for enabling all stages of architecture and urban design.

%%%%%%%%%%%%%%%%%%%%%%%%%%%%%%%%%%%%%%%%%%%%%%%%%%%%%%%%%%%%%%%%%%%%%%%%%%%%%%%%

\begin{acks}
This work is funded in part by the National Natural Science Foundation of China Projects No. U1936213, No.62176185. This work is also partially supported by the Shanghai Science and Technology Development Fund No. 19DZ1200802, and by the Shanghai Municipal Science and Technology Major Project (2021SHZDZX0100) and the Fundamental Research Funds for the Central Universities.
\end{acks}

\bibliographystyle{ACM-Reference-Format}
\balance
\bibliography{sample-base}

\appendix

% \section{ReCo Dataset Statistics}

% \input{ins_stas}
% \input{inst_stas_detail}

\section{Details of Data Instance}
\label{app_instance}
\begin{table}[b]
    \centering
    \setlength{\tabcolsep}{0.01mm}{
        \begin{tabular}{@{}cccc@{}}
            \toprule
            Instance                   & Data         & Data Format & Example/ Describe                                                              \\ \midrule
            \multirow{1}{*}{City} & Community         & Instances      & A set of Community instances   \\ \midrule
            \multirow{4}{*}{Community} & \_id         & String      & ``61ef8a8b32b5d4672152cf73''                                                    \\
                           & boundary     & Coordinates & \begin{tabular}[c]{@{}l@{}}A set of 2D coordinates\end{tabular} \\
                           & Building     & Instances   & A set of Building instances                                                   \\
                           & City         & String      & ``city\_16''                                                                    \\ \midrule
            \multirow{3}{*}{Building}  & building\_id & Int         & 35710                                                                         \\
                           & floor        & Int         & 3                                                                             \\
                           & coords       & Coordinates & \begin{tabular}[c]{@{}l@{}}A set of 2D coordinates\end{tabular} \\ \bottomrule 
        \end{tabular}
        }
        \caption{Details of each data instance.}
\end{table}

\end{document}